%% file: main.tex
\documentclass[runningheads]{llncs}

\usepackage{eccv}

\usepackage{eccvabbrv}

\usepackage{graphicx}
\usepackage{booktabs}

\usepackage[accsupp]{axessibility}  

\usepackage[pagebackref,breaklinks,colorlinks,citecolor=eccvblue]{hyperref}

\usepackage{orcidlink}

\begin{document}

\title{QUASAR: QUality and Aesthetics Scoring with Advanced Representations}

\titlerunning{QUASAR: Quality and Aesthetics}

\author{Sergey Kastryulin\inst{1,3\star}
\and
Denis Prokopenko\inst{2}\thanks{S.K. \& D.P. contributed equally.}
\and
Artem Babenko\inst{3}
\and\\
Dmitry V. Dylov\inst{1,4}
}

\authorrunning{S.~Kastryulin \textit{et al.}}

\institute{Skolkovo Institute of Science and Technology \and
King's College London
\and
Yandex
\and
AIRI}

\maketitle

\input{ECCV/chapters/00_abstract}

\input{ECCV/chapters/01_intro}

\input{ECCV/chapters/02_related}

\input{ECCV/chapters/03_method}

\input{ECCV/chapters/04_experiments}
\input{ECCV/chapters/13_discussion}


\input{ECCV/chapters/10_conclusion}

\input{ECCV/chapters/_supplementary}


\clearpage  

%
%
\bibliographystyle{splncs04}
\bibliography{main}
\end{document}

%% file: ECCV/chapters/00_abstract.tex
\begin{abstract}



This paper introduces a new data-driven, non-parametric method for image quality and aesthetics assessment, surpassing existing approaches and requiring no prompt engineering or fine-tuning. 
We eliminate the need for expressive textual embeddings by proposing efficient image anchors in the data.
Through extensive evaluations of 7 state-of-the-art self-supervised models, our method demonstrates superior performance and robustness across various datasets and benchmarks.
Notably, it achieves high agreement with human assessments even with limited data and shows high robustness to the nature of data and their pre-processing pipeline. 
Our contributions offer a streamlined solution for assessment of images while providing insights into the perception of visual information.
\keywords{Image Quality Assessment \and Foundation Models \and Self-supervised learning \and Metrics and Benchmarks \and Multi-modal Representations}

\end{abstract}

%% file: ECCV/chapters/01_intro.tex
\section{Introduction}
\label{sec:intro}


In recent years, image quality assessment (IQA) and image aesthetics assessment (IAA) have gained significant attention as crucial components of various computer vision tasks, ranging from image recovery, to image enhancement, to image generation, to multimedia content sharing \cite{sara2019iqa_survey}. 
However, capturing the essence of image quality and that of the aesthetics are highly challenging tasks, given the intricacies of the human perception and the subjectivity \cite{sara2019iqa_survey}. 

In this context, the emergence of self-supervised learning methods, such as the CLIP image-to-text model \cite{radford2021learning}, offers a promising avenue for creating powerful image assessment tools. 
CLIP is found to produce highly descriptive image and text embedding, applicable well-beyond their original purpose of image search and retrieval.
Lately, the CLIP-IQA model \cite{wang2023exploring} was first to tailor the use of textual descriptors, such as ''\textit{Bad image.}'' and ``\textit{Good image.}'', to the task of quality assessment.

\begin{figure}[tb]
  \centering
  \includegraphics[width=0.9\textwidth]{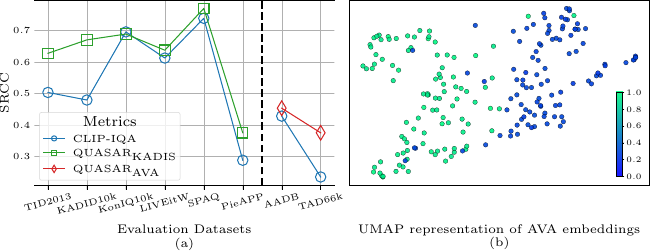}
  \caption{
    \textbf{(a)} The best performance of QUASAR and CLIP-IQA metrics on 6 image quality and 2 image aesthetics benchmarks. 
    Note the large gain on widely used TID2013, KADID10k, and the most challenging TAD66k datasets. 
    \textbf{(b)} The proposed metric enables a clear separation of embeddings, as shown for the AVA data.  
  }
  \label{fig:best}
\end{figure}

In our work, we expand the scope of this approach using a standpoint of a unified score, where one could gauge not only the technical quality of an image but also its aesthetic value in a single run.
Unlike the majority of modern research efforts, we propose to do that \textit{not by the prompt-engineering}, but via a new data-driven approach of non-parametric assessment of \textit{image characteristics}. 
In the proposed paradigm, the technical, content-based, and the perceptual image qualities are deemed as generic characteristics responsible for the subjective opinion about an image. 
We show that this boosts the performance and holds the potential to bridge the gap between parametric and non-parametric methods in the field of image assessment.

\textbf{The scope of our study.}
We explore \textit{general-purpose} foundation models which were never purposely trained for the image assessment. This is different from standard approaches that either directly train on the mean opinion scores (\textit{e.g., PieAPP} \cite{prashnani2018pieapp}, such metrics as MANIQA\cite{yang2022maniqa} or DISTS \cite{ding2020image}), or indirectly train on the types of distortions and their strengths (\textit{e.g., CONTRIQUE} \cite{madhusudana2022image}). The reason behind this decision is that we purposely want to avoid the need for the distortion data or the meta-data, typically entailed in devising new IQA metrics. This excludes both reference and no-reference IQA approaches in our comparison. Instead, our motivation is to find out if the foundation models could be used as proxies for assessing images.

Hence, in our study we delve deeply into an extensive analysis of 7 modern self-supervised model architectures in combination with 3 types of anchor features and validate the proposed score on 8 various datasets and benchmarks (Refer to Fig. \ref{fig:best} for a summary). 
We also examine the important question of data requirements and conclude that high agreement with the human assessments can be achieved even with a limited set of samples.


Our contributions are the following:
\begin{itemize}
    \item We propose a new data-driven method for non-parametric assessment of image characteristics (quality and aesthetics), which does not require prompt-engineering and significantly improves previously reported results.
    \item We provide an extensive analysis of 7 state-of-the-art self-supervised approaches and model architectures, showcasing that our approach significantly outperforms previous non-parametric image quality assessment methods in terms of both peak performance without fine-tuning and robustness across different datasets and benchmarks.
\end{itemize}

%% file: ECCV/chapters/02_related.tex
\section{Background and Related Work}
\label{sec:related}

\textbf{Image quality and aesthetics assessment} aspires to quantify the appeal of an image, taking into account both its technical quality (\textit{e.g.}, presence of blur or noise) and its artistic value (\textit{e.g.}, composition or style). 

Initial assessment methods were based on human perception~\cite{wang2006modern}, scene statistics~\cite{sheikh2005visual,mittal2012no}, or similarity~\cite{zhang2011fsim} with or without the reference images; however, these methods were limited by our understanding of the human vision system and scene interpretation.
Newer approaches~\cite{prashnani2018pieapp,cheon2021perceptual,lao2022attentions} relied on fitting to TID2013~\cite{ponomarenko2015image}, KADID10k~\cite{Lin2019kadid10k}, PIPAL~\cite{jinjin2020pipal}, and other IQA datasets with opinion scores provided by humans.
Recent advances in multi-modal networks enabled new, so-called \textit{non-parametric}, methods~\cite{wang2023exploring}. 
These methods compromise optimal performance for greater generalizability, making new metrics closer to the real-world application scenarios.
Herein, we build up on the research outcomes in this area aiming to extend and improve state-of-the-art results.

\textbf{Self-supervised Models}, such as CLIP \cite{radford2021learning}, ALIGN \cite{jia2021align}, DINO \cite{caron2021emerging},  DINOv2 \cite{oquab2023dinov2}, BLIP \cite{li2022blip}, BLIP-2 \cite{li2023blip2},  COCA \cite{yu2022coca}, and others are known for their extensive use beyond the scope of what they were originally trained for.
Generic features learned by these models could be applied in various image understanding tasks, including zero-shot image classification \cite{radford2021learning}, object detection \cite{lin2023gridclip}, semantic segmentation \cite{zhou2023zegclip}.
Recently, CLIP became a vital component of generative text-to-image models, such as Stable Diffusion \cite{rombach2021highresolution}, enabling the production of high-quality and aesthetically pleasant images.
For that to be possible, CLIP has had to be sensitive enough to the fine-grained multi-modal characteristics of the data, which was partially supported by the recent CLIP-IQA model \cite{wang2023exploring}.
Yet, we argue that the IQA potential of self-supervised models is not yet harnessed to its fullest and extend their applicability to other types of non-semantic image characteristics, such as image aesthetics.

Some methods not only use pre-trained self-supervised models but also fine-tune them (CLIP-IQA+ \cite{wang2023exploring}) or provide trainable adaptors for them (AesCLIP \cite{sheng2023aesclip}).
Despite showing promising results in specific tasks, these methods compromise generalizability: arguably, any task-specific training compromises generalizability of a method, which is why we focused on a \textit{completely non-parametric} approach.

\textbf{Embeddings expressiveness} is a vital and a sought-after property of self-supervised models.
The majority of self-supervised models considered are \textit{multi-modal}, operating with both image and text data.
``\textit{An image is worth a thousand worlds}'', however, which refers to the fact that natural language can be vague and more ambiguous than a visual expression of the same thing. 
This was a motivation of the authors of unCLIP \cite{ramesh2022unclip} text-to-image diffusion model.
They designed ``a prior block'' to map CLIP text embeddings to CLIP image embeddings before transmitting them to the model for conditional generation.
Our method is motivated by that same idea.
By proposing \textit{image anchors} over \textit{prompt pairs} we aim to define the notion of positive and negative samples for the unified assessment of quality and aesthetics more firmly.

%% file: ECCV/chapters/03_method.tex
\section{QUASAR}
\label{sec:method}
\input{ECCV/figs/quasar_main}


We delineate the methodology employed to develop a new data-driven approach for non-parametric assessment of image quality and aesthetics in Fig. \ref{fig:quasar_main}. 
The proposed method is called QUASAR (QUality and Aesthetics Scoring with Advanced Representations), with its ingredients described separately below.


\subsection{Anchor Data}



Anchor data serves as a reference point to which the image representations are compared, allowing for a meaningful quantification of image quality and aesthetics on a relative scale. 
The anchor data could consist of handcrafted or automatically learned features that represent desirable or undesirable image characteristics, such as sharpness, noise, color balance, or composition. 
By capturing various aspects of image quality and aesthetics, the anchor data facilitates the evaluation process by providing a reference for comparison.

There are several ways to define anchor data. 
The CLIP-IQA \cite{wang2023exploring} approach previously suggested using text pairs like \textit{"Good image."} and \textit{"Bad image."} as anchors. 
However, employing prompts may not be the most effective strategy since it introduces the inherent ambiguity associated with natural languages. 
As a result, we propose utilizing combinations of image embeddings extracted using various image encoders as anchor data. 
This approach enables the construction of a more fine-grained and expressive reference space, thereby improving the robustness of our method in assessing image quality.

Using embeddings as anchor data has several advantages. 
First, it allows for a more direct comparison between image representations and reference points, avoiding potential issues introduced by the varied interpretations of textual prompts. 
Second, it enables the exploitation of the full expressive power of the image encoder, as the anchor data is represented in the same feature space as the input images. 
Finally, leveraging embeddings as anchor data fosters a data-driven approach, as the reference points can be obtained from a diverse set of images with corresponding quality or aesthetic labels, thus contributing to better overall quality in the assessment process.

\subsection{Image Encoder}


The primary function of the Image Encoder is to map images into a high-dimensional feature space, where the resulting embeddings capture discriminative and informative properties of the images. 
These characteristics may align with image quality or aesthetics, allowing for effective evaluation of visual content. 

Our approach is flexible in terms of the Image Encoder utilized, as we consider a variety of state-of-the-art self-supervised models, including CLIP \cite{radford2021learning}, DINOv2 \cite{oquab2023dinov2}, and COCA \cite{yu2022coca}.
For each model we consider several backbones, including ViTs \cite{dosovitskiy2020image} and ResNets \cite{jian2016deep}.
By examining the performance of these models, we aim to identify the extractors that are most sensitive to low-level image properties and provide a thorough understanding of the features that contribute to the perception of image quality and aesthetics.

\subsection{Aggregation Function}


The aggregation function takes the image embeddings and anchor data as input and computes a centroid vector that reflects the quality or aesthetics of the Anchor Data. 
The function is designed to be non-parametric and free from any specific assumptions, allowing for a flexible and robust assessment of image properties across different datasets and benchmarks.

Assuming uneven distribution of samples the aesthetic data, we consider the following aggregation strategies: \textit{Mean} averaging of all embeddings in each subset, mean averaging of a fraction of each subset defined by an \textit{Offset} from the division point and K-means \textit{Clustering} \cite{lloyd1982least} on $N$ clusters with averaging of inter-cluster samples and subsequent averaging of cluster centroids. 

The primary purpose of the Aggregation Function is to integrate the information captured by the representation extractor and the anchor data and produce a single, informative embedding. 
This embedding is further analyzed by determining the extent to which it aligns with the input image of interest.
The degree of alignment is expected to be related to the human perception of image quality or aesthetics.

\subsection{Score Computation}

Having two centroids obtained from the Aggregation Function as anchors representing images with high and low quality/aesthetics, we are ready to compute the final  score $\bar{s}$.
Given an input image, we extract its embedding using the same Image Encoder.
Let $c_1$ and $c_2$ be the centroids that define the notion of images with high and low quality/aesthetics and $x$ be the embedding of the input image, we first compute the cosine similarity

\begin{equation}
\label{eq:cosine_sim}
s_i = \frac{x \odot c_i}{||x|| \cdot ||c_i||}
\end{equation}

We compute softmax to obtain the final score

\begin{equation}
\label{eq:softmax}
\bar{s} = \frac{e^{s_1}}{e^{s_1} + e^{s_2}}
\end{equation}

We argue that the softmax is not the only way of getting final scores from similarity estimations but we leave a more detailed investigation of that matter for future work. 

The usage of a pair of centroids helps to avoid the ambiguity as the task is now cast as a binary classification, where the final score is regarded as a relative similarity as done in \cite{wang2023exploring}.
We estimate the correctness of the obtained scores by computing Spearman’s Rank-order correlation coefficient (SRCC) with labelers’ mean opinion scores.

\begin{equation}
\label{eq:srcc}
\text{SRCC} = 1 -  \frac{6 \sum_{i=1}^{n} d_i ^ 2}{n (n^2 - 1)},
\end{equation}
\noindent
where $d_i$ is the difference between the \textit{i}-th image’s ranks in the objective and the subjective ratings and $n$ is the number of observations.

Below we discuss our experimental results and report the performance of all metrics in terms of SRCC, which we consider to be the ultimate measure of metrics' performance.

\input{ECCV/tables/results}

%% file: ECCV/figs/quasar_main.tex
\begin{figure*}[tp]
    \centering
    \includegraphics[width=0.95\linewidth]{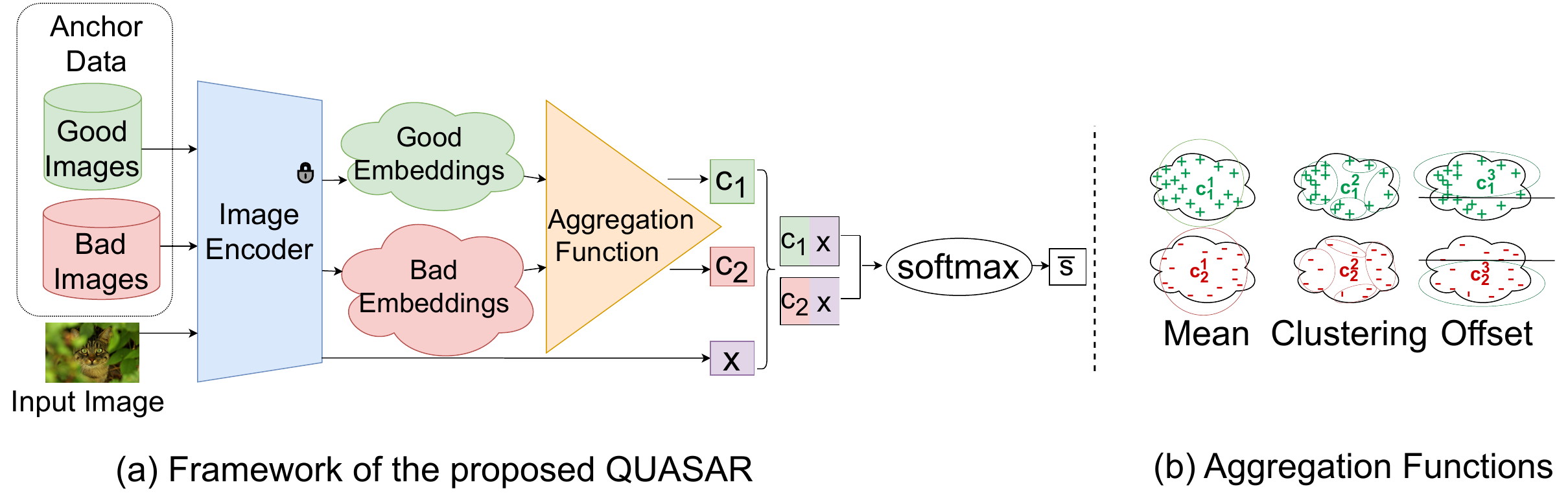}
    \caption{(a) QUASAR framework consists of Anchor Data, Image Encoder, and Aggregation Function. 
    Geven an input image, the centroids are used to compute the final score $\bar{s}$.
    (b) Three types of Aggregation Functions. 
    Each Aggregation Function produces centroids with distinct properties.}
    \label{fig:quasar_main}
\end{figure*}

%% file: ECCV/tables/results.tex
\begin{table*}[tbh!]
\caption{Comparison of 8 popular IQA and IAA metrics, as well as recently proposed CLIP-IQA, with the best QUASAR variant. By * we denote the results for \textit{learning-based} methods, included as a reference rather than for a direct comparison. 
Best performer on each dataset is marked in \textbf{bold}.}
\vspace{3mm}
\setlength\tabcolsep{2.5pt}
\begin{center}
\footnotesize{
\resizebox{1\columnwidth}{!}{ 
\begin{tabular}{lcccccccclcc}
\toprule
&\multicolumn{7}{c}{\textbf{Image Quality}}&&\multicolumn{3}{c}{\hphantom{AAAAAAA}\textbf{Aesthetics}}  \\
\cmidrule{2-7} \cmidrule{11-12}
& TID2013 & KADID10k & KonIQ10k & LIVEitW & SPAQ & PieAPP && \hphantom{} && AADB & TAD66k \\

  \cmidrule{2-7} \cmidrule{11-12}
  BIQI \cite{moorthy2010two} & $0.393$ & $0.338$ & $0.559$ & $0.364$ & $0.591$ & $0.243$ && &
  NIMA \cite{talebi2018nima} & $0.612$* & $0.390$* \\
  BLINDS-II \cite{saad2012blind} & $0.393$ & $0.308$ & $0.585$ & $0.090$ & $0.317$  &  $0.250$ && &
  ALamp \cite{ma2017lamp} & $0.666$* & $0.411$* \\
  BRISQUE \cite{mittal2012no} & $0.370$ & $0.332$ & $0.705$ & $0.561$ & $0.484$ & $0.328$ && & 
  RAPID \cite{lu2014rapid} & $0.447$* & $0.314$* \\
  NIQE \cite{mittal2012making} & $0.313$ & $0.379$ & $0.551$ & $0.463$ & $0.703$ & $0.182$ && &
  AADB \cite{kong2016photo} & $0.558$* & $0.379$* \\
\cmidrule{1-12}

  CLIP-IQA \cite{wang2023exploring} & $0.503$ & $0.479$ & $\textbf{0.695}$ & $0.612$ & $0.738$ & $ 0.288$ && & \hphantom{AAA} & $0.428$ & $0.235$ \\
  QUASAR (ours) & $\textbf{0.627}$ & $\textbf{0.670}$ & $0.688$ & $\textbf{0.637}$ & $\textbf{0.769}$ & $\textbf{0.375}$ && & \hphantom{AAA} & $\textbf{0.453}$ & $\textbf{0.375}$ \\

\bottomrule
\end{tabular}
}}
\end{center}
\label{table:results}
\end{table*}

%% file: ECCV/chapters/04_experiments.tex
\section{Experiments}
\label{sec:experiments}




\input{ECCV/figs/supp_1_tops_LIVEitw_only}

We use two pools of data to form the \textbf{Anchor Data}: KADIS700k dataset \cite{Lin2019kadid10k} for IQA and AVA dataset \cite{murray2012ava} for IAA.

KADIS700k is a full-reference dataset, meaning that all distorted images are obtained from the reference ones by corrupting them with a predefined set of distortions.
This way of dataset collection results in a natural division into two subsets, namely reference and distorted data, representing images of high and low image quality.

IAA datasets do not have such a natural division so we split the employed AVA in two equally sized parts by median opinion score value.
We hypothesize that the same procedure could be utilized for division of no-reference IQA datasets to obtain alternative, more realistic anchor data for quality assessment. 
We leave the verification of this hypothesis for future work.

In our work we rely on OpenCLIP implementations \cite{ilharco_gabriel_2021_5143773,cherti2023reproducible} of  CLIP \cite{radford2021learning}, DINOv2 \cite{oquab2023dinov2}, and COCA \cite{yu2022coca}.
The general pre-processing procedure for these models include: i) resize with bicubic interpolation to at least one spatial dimension resolution of 224 px, ii) central crop to 224 x 224 px RGB image, iii) $[0, 1]$ normalization and standardisation to ImageNet \cite{deng2009imagenet} mean and standard deviation.

However, CLIP-IQA \cite{wang2023exploring}, our main baseline, uses a peculiar modification of the above pipeline. 
The resizing step is skipped for all datasets except from SPAQ \cite{fang2020perceptual}, where the images are resized to one resolution but to 512 px instead of standard 224 px.
We stick with no resizing for IQA for a fair comparison while using the vanilla OpenCLIP pre-processing pipeline for aesthetics.
Below we ablate the choice of resizing strategy, revealing its major impact on metrics' performance.

\textbf{Extraction and Aggregation.}
For the best results we use CLIP-RN50 with no positional embeddings for IQA and CLIP-ViT-L/14 without any changes for IAA.
We use both backbones to extract all image embeddings from both anchor and evaluation data.

We divide AVA dataset by median mean opinion score value and aggregate its parts using k-means clustering \cite{lloyd1982least}. 
For that, we first find centroids of 100 clusters for each subset by simple averaging intra-cluster embedding values.
Second, we find centroids of both anchor data subsets by averaging out the cluster centroids. 

\subsection{Results}
\label{sec:results}

Following the standard evaluation procedure, we compute correlation values in terms of SRCC score (Eq. \ref{eq:srcc}) with mean opinion scores from human assessors on 7 benchmark datasets, including 3 full-reference IQA datasets (KADID10k \cite{Lin2019kadid10k}, TID2013 \cite{he2022rethinking}, PieAPP \cite{prashnani2018pieapp}), 3 no-reference IQA datasets (KonIQ10k \cite{hosu2020koniq10k}, LIVEitW \cite{ghadiyaram2015massive}, SPAQ  \cite{fang2020perceptual}), and 2 IAA datasets (AADB \cite{kong2016photo}, TAD66k \cite{he2022rethinking}).

\input{ECCV/figs/experiments_2_backbones}

We compare QUASAR with alternative non-learning based methods for no-reference IQA.
While all relatively new IAA metrics are learning-based, it does not stop us from referring them, keeping in mind their clear competitive advantage. 
Our main reference is the recently proposed CLIP-IQA \cite{wang2023exploring} metric, which is, like QUASAR, a generic approach, applicable for assessment of both image quality and aesthetics.

Table \ref{table:results} shows that QUASAR surpasses all relevant IQA baselines and performs similar to learning-based IAA methods without any need of model training or fine-tuning.  
Moreover, we show that the proposed approach is noticeably more robust than previously proposed CLIP-IQA model without task- and dataset-specific pre-training or prompt tuning (Figure \ref{fig:last_resol}).
In particular, we draw the reader's attention on QUASAR's generalizability for the most challenging IAA dataset TAD66k and full-references IQA datasets (TID2013, KADID10k), which pose a great interest with their rich collection of distortions.
Refer to Fig. \ref{sup:scores_liveitw} and the Supplementary material for visual assessment of the image metrics, sorted according to the MOS.

\subsection{Ablation Study}

Below we discuss various factors influence performance and robustness of both CLIP-IQA and QUASAR frameworks.

\subsubsection{Feature Extractors and Resizing}

While QUASAR is able to achieve robust results across several benchmarks, we aim to investigate influence of the backbone and resizing strategy to see its pros and cons compared to CLIP-IQA.
For that, we consider alternative self-supervised learning frameworks such as DINOv2 \cite{oquab2023dinov2} and COCA \cite{yu2022coca} together with a set of Visual Transformer \cite{dosovitskiy2020image} backbones of various sizes.
\input{ECCV/figs/experiments_last_resolutions}

It was previously reported \cite{wang2023exploring} that among all backbones, ResNets are the least sensitive to the removal or modification of positional embeddings. 
That was the deciding factor for their usage in the CLIP-IQA framework with two main strategies regarding positional embeddings: their complete removal and interpolation to adjust for different input image sizes. 
At the same time, Visual Transformers are more sensitive to manipulations with positional embeddings, which led to noticeable performance degradation. 

As an alternative, one could use the standard OpenCLIP pre-processing procedure that includes resize of all input images to \textit{224 px} resolution. 
In that case there will be no need to modify positional embeddings in any of the above models.
Figure \ref{fig:backbones} shows that this modification  effects CLIP-IQA and QUASAR performance in different scales.
In general, the gap between the two methods increases regardless of IQA Anchor data used (discussed below). 
We argue that this observation indicates high generalizability of our approach, showing that even significant changes in the pre-processing pipeline lead to virtually no performance changes.

\input{ECCV/figs/experiments_3_ratios}

\subsubsection{Anchor Data and Their Size}
\label{sec:abl-data-size}

QUASAR framework heavily relies on Anchor Data as a source of information regarding image quality and aesthetics.
In \ref{sec:results} we discussed the best results obtained with KADIS700k and AVA datasets as anchor data, containing 840,000 and 255,000 images respectively.
This fact immediately raises the following questions: i) how the particular choice of the Anchor Data influences the performance? and ii) is it required to have such a large dataset to obtain a high-quality metric?

To address the first question we consider PIPAL \cite{jinjin2020pipal} as an alternative IQA dataset.
PIPAL differs from KADIS700k in terms of its size and content.
It consists of 29,000 images, most of which are obtained from 250 reference ones by adding specifically designed IQA distortions. 
We used only the publicly available train subset of the PIPAL dataset, resulting in 200 reference and 23,200 distorted images.
KADIS700k uses different set of distortions and has greater fraction of clean reference images (140,000 out of 840,000).
Nevertheless, Fig. \ref{fig:backbones} shows that even in  case of significant change of the Anchor Data QUASAR demonstrates decent results, sometimes even outperforming the initially proposed KADIS700k version.

To address the dataset size question, we take random subsets of Anchor Datasets to form the centroids.
Figure \ref{fig:ratio_linked} shows rapid saturation of metric's quality with number of samples used as Anchor Data suggesting that only little fraction of the used datasets may be enough to achieve best results.
Moreover, large standard deviation in low fraction regions suggests that there might exist small subsets consisting of no more than several images that might deliver competitive quality.

\input{ECCV/figs/experiments_567_combined}

The same effect can be observed for the aesthetics datasets.
Figure \ref{fig:offest_aggregation_CLIP} indicates that for both TAD66k and AADB the use of only 30\% of datasets for centroids generation leads to better performance when using the \textit{Mean} aggregation.
Such a behaviour evidences that there might be a more informative, non-random subset of images that better captions the notion of high and low aesthetic attractiveness.
This observation motivates us to investigate the use of different aggregation functions.

\subsubsection{Aggregation Function}

In case of IAA, we do not have a strict reference-distorted distinction between images in contrast to IQA datasets, such as PIPAL and KADIS700k.
Therefore, we need to adjust the data split to deliver a better performance. 

We consider 3 aggregation strategies. 
\textit{The first one} is a simple averaging of all embeddings in each subset, we call it (\textit{Mean}) aggregation.
\textit{Second}, we extend the \textit{Mean} method by the use of an offset from the division point.
The analysis of the AVA dataset \cite{he2022rethinking} shows that the distribution of mean opinion scores is extremely uneven and the majority of samples have score values close to median.
Large variations of image content on images with similar aesthetic scores may play a role of noisy data.
We experiment with the \textit{Offset} method to mitigate this effect.
\textit{In the third strategy}, we experiment with \textit{k}-means \textit{Clustering} \cite{lloyd1982least} on $N$ clusters with averaging of inter-cluster samples and subsequent averaging of cluster centroids. 
We hypothesize that if data contain many similar images in terms of their scores and content, these images may pull centroids off while valuable rare samples will almost be ignored.
The main motivation of this method is to re-weight samples to mitigate such distribution imbalances.

\input{ECCV/figs/supp_3_examples_1}

Figure \ref{fig:offest_aggregation_CLIP}(c) gives an overview of the performance of the selected aggregation methods. 
Maximal performance with medium number of clusters and a peak for high offset values on AADB dataset suggest that there is only a fraction of informative samples with regard to the aesthetics of the images.
This observation, in conjunction with the one from Section \ref{sec:abl-data-size}, suggest an avenue for further investigation of which images tend to be the most informative and why.

Figure \ref{sup:kadid10k_distortion} clearly demonstrates that, when comparing QUASAR and CLIP-IQA scores on the KADID10k samples, both QUASAR and CLIP-IQA exhibit similar trends across different types and levels of distortions. However, QUASAR scores demonstrate a wider range of scores compared to CLIP-IQA, making it a more sensitive choice for an image metric.

%% file: ECCV/figs/supp_1_tops_LIVEitw_only.tex
\begin{figure*}[tp]
    \centering
    \begin{subfigure}{0.99\textwidth}
        \centering
    
        \includegraphics[width=0.9\textwidth]{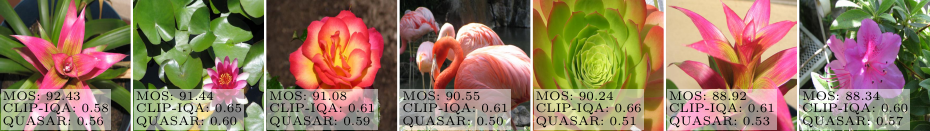}
        \caption{}
    \end{subfigure}
    \\
    \begin{subfigure}{0.99\textwidth}
        \centering
    
        \includegraphics[width=0.9\textwidth]{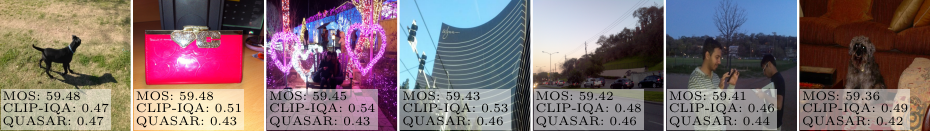}
        \caption{}
    \end{subfigure}
    \\
    \begin{subfigure}{0.99\textwidth}
        \centering
    
        \includegraphics[width=0.9\textwidth]{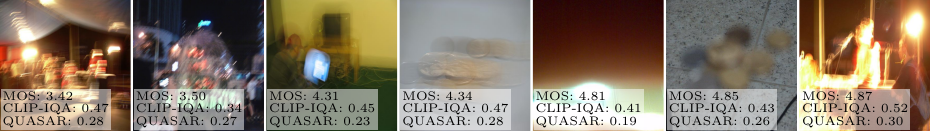}
        \caption{}
    \end{subfigure}
    
    \caption{Samples from LIVEitW dataset, divided into three categories, according to MOS: (a) top, (b) median, and (c) low.}
    \label{sup:scores_liveitw}
\end{figure*}

%% file: ECCV/figs/experiments_2_backbones.tex
\begin{figure}[tp]
    \centering
    \includegraphics[width=0.9\linewidth]{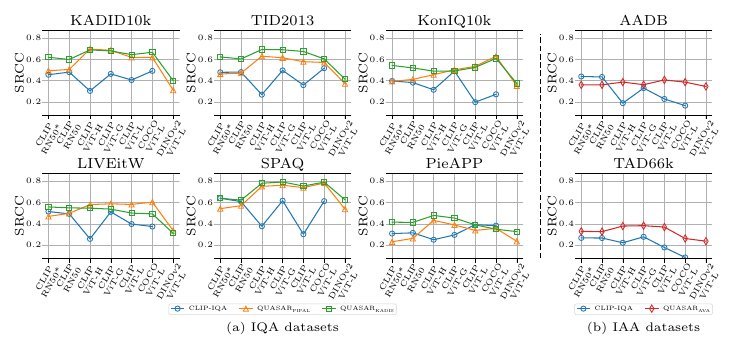}
    \caption{Comparison of QUASAR and CLIP-IQA variants with standard OpenCLIP \cite{ilharco_gabriel_2021_5143773,cherti2023reproducible}, resizing to \textit{224 px} resolution and allowing for direct use of ViT backbones.
    RN50 and RN50* denote ResNet-50 backbones with and without positional embeddings, respectively.
    Note the gain in performance between CLIP-IQA and QUASAR, showcasing robustness of the latter to small tweaks in the pre-processing pipeline. 
    (a) Variants of QUASAR for IQA with varying the anchor data (PIPAL, KADIS700k). 
    Despite being smaller and having different set of distortions, PIPAL version has comparable performance, showcasing robustness of QUASAR to the choise of anchor data. 
    (b) QUASAR variant with \textit{Mean} aggregation and resizing on IAA datasets. 
    Note the large and consistent performance gain over the prompt-based metric.}
    \label{fig:backbones}
\end{figure}

%% file: ECCV/figs/experiments_last_resolutions.tex
\begin{figure}[tp]
    \centering
   \includegraphics[width=0.9\linewidth]{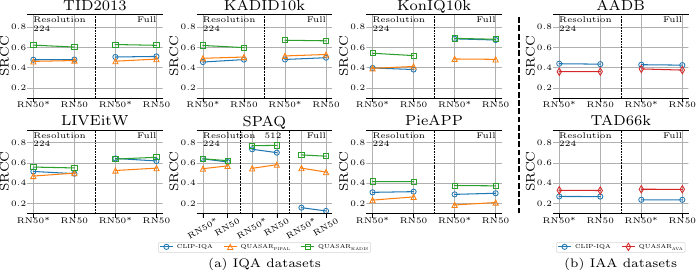}
    \caption{QUASAR and CLIP-IQA performance, depending on the resolution of input data.
    Unlike QUASAR, CLIP-IQA shows a significant drop of SRCC when images are resized to the uniform \textit{224 px} resolution on SPAQ, KonIQ10k, and LIVEitW datasets. 
    RN50 and RN50* denote ResNet-50 backbones with and without positional embeddings, respectively.
    }
    \label{fig:last_resol}
\end{figure}

%% file: ECCV/figs/experiments_3_ratios.tex
\begin{figure}[tp]
    \centering
    \includegraphics[width=0.9\linewidth]{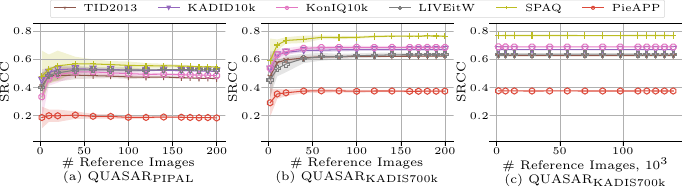}
    \caption{Relation between the fraction of Anchor Data used to form the centroids and the resulting QUASAR performance.
    Note that the scale of KADIS700k dataset (consists of 840k images in total) makes the performance virtually constant (c). 
    Figures (a) and (b) consider regimes with lower number of samples and suggest that high performance can we achieved with as many as just a few samples (both with PIPAL and KADIS700k Anchor Data).}
    \label{fig:ratio_linked}
\end{figure}

%% file: ECCV/figs/experiments_567_combined.tex
\begin{figure}[tp]
    \centering
    \includegraphics[width=0.9\linewidth]{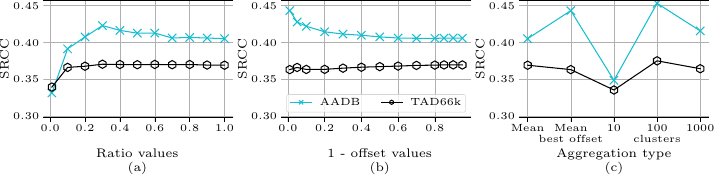}
    \caption{Aesthetics ablation with CLIP ViT-L14. (a) Random subsampling. (b) Offset according to MOS scores. (c) Aggregation methods.}
    \label{fig:offest_aggregation_CLIP}
\end{figure}

%% file: ECCV/figs/supp_3_examples_1.tex
\begin{figure*}[tp]
    \centering
    \begin{subfigure}{0.49\textwidth}
        \includegraphics[width=\textwidth]{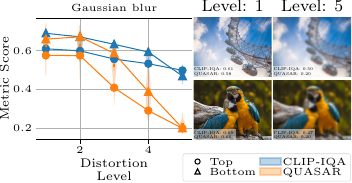}
        \caption{}
    \end{subfigure}
    \hfill
    \begin{subfigure}{0.49\textwidth}
        \includegraphics[width=\textwidth]{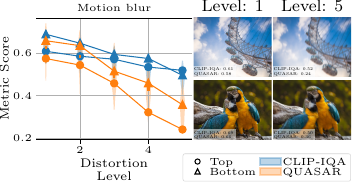}
        \caption{}
    \end{subfigure}
    \\
    \begin{subfigure}{0.49\textwidth}
        \includegraphics[width=\textwidth]{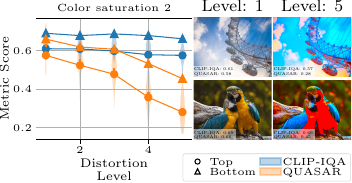}
        \caption{}
    \end{subfigure}
    \hfill
    \begin{subfigure}{0.49\textwidth}
        \includegraphics[width=\textwidth]{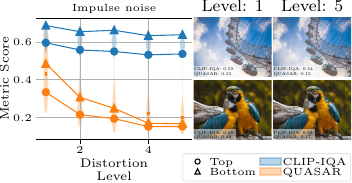}
        \caption{}
    \end{subfigure}
    \\
    \begin{subfigure}{0.49\textwidth}
        \includegraphics[width=\textwidth]{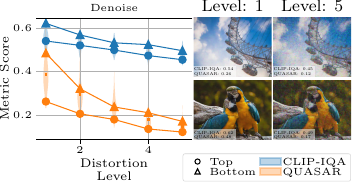}
        \caption{}
    \end{subfigure}
    \hfill
    \begin{subfigure}{0.49\textwidth}
        \includegraphics[width=\textwidth]{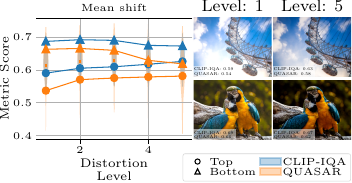}
        \caption{}
    \end{subfigure}
    \\
    \begin{subfigure}{0.49\textwidth}
        \includegraphics[width=\textwidth]{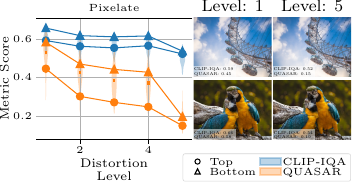}
        \caption{}
    \end{subfigure}
    \hfill
    \begin{subfigure}{0.49\textwidth}
        \includegraphics[width=\textwidth]{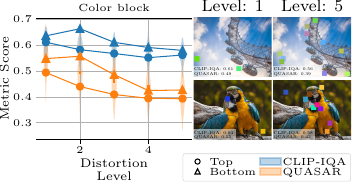}
        \caption{}
    \end{subfigure}

    \caption{Comparison of QUASAR and CLIP-IQA scores on samples from KADID10k dataset for different levels of applied distortions: (a) Gaussian blur, (b) Motion blur, (c) Color saturation 2, (d) Impulse noise, (e) Image denoising, (f) Mean shift, (g) Pixelate, and (h) Color block. Both trends in QUASAR and in CLIP-IQA match across the distortion types and levels, but QUASAR scores have a higher dynamic range. Raw images are shown, while both QUASAR and CLIP-IQA take advantage of data preprocessing, as described in the main text. The evaluated scores are normalised to $[0,1]$ range.}
    \label{sup:kadid10k_distortion}
\end{figure*}

%% file: ECCV/chapters/13_discussion.tex
\section{Discussion}
\label{sec:discussion}
We introduced a new data-driven method for non-parametric image quality and aesthetic assessment, surpassing other non-parametric methods without the need for prompt-engineering. Comparison with 7 top self-supervised techniques and model architectures confirms the superior performance of our approach in joint image quality and aesthetics assessment.
Technical and quantitative advantage of our work over pre-existing CLIP-based methods was thoroughly shown, including an exhaustive quantitative (Table \ref{table:results}, 
Figs. \ref{fig:backbones} \& \ref{fig:last_resol}) and qualitative (Fig. \ref{sup:scores_liveitw} and Supplementary material) comparison with the CLIP-based baselines. 
The gain is obvious in terms of both peak performance without any fine-tuning and the metrics generalizability, showing robustness across different datasets and benchmarks.

Note that, in our results, the SRCC score was reported as a measure of agreement with the user opinion (MOS); we consider this score the most relevant for our study. We find Pearson's linear correlation coefficient to be less descriptive because MOS and the metrics' values are typically non-linearly correlated (refer to \cite{our_iqa_for_mri_paper}, also confirmed herein empirically). Conversely, KRCC is highly-correlated with SRCC, as they both correspond to the same (rank order) family of methods. 
We also found the UMAP visualization, shown in Fig. \ref{fig:best}, useful for gauging the expressiveness of the embeddings based on the proposed anchor data. The UMAP clearly shows the separate clusters and confirm our indirect observations in Fig. \ref{fig:offest_aggregation_CLIP} (b) of the ablation study.

The main conclusion of the ablation and the anchor data types analysis is the vital importance of the choice of the anchor data. 
We performed additional experiments where a small subset of anchor data was \textit{altered adaptively} based on the properties of the test samples (using, for instance, the retrieval-augmented paradigm). Remarkably, in such a way, the performance of QUASAR could be boosted even higher. We obtain SRCC of up to 0.859 on AVA dataset with TAD66k embeddings as anchors and suggest a more systematic study of this idea for the future work.  

Following our study of \textit{general-purpose} foundation models that were never purposely trained for the image assessment, we can see that one can indeed devise an efficient image assessment metric (Refer to Fig. \ref{fig:best} for a summary). The advantage is evident: no need to manually generate the distortion data or the meta-data, typically entailed in devising new IQA metrics. The foundation model already `knows' the average preferences of the user. Anticipating a natural critique that these foundation models require a lot of data, we can say that our experiments suggest that a high agreement with the human assessments can be achieved even with a limited set of samples (See Fig. \ref{fig:ratio_linked}).
Moreover, the rapid development of large foundation models is already a given and occurs independently; hence, their use as proxies for the image assessment is sound and justified regardless of the data demands.

Another noteworthy advantage of harnessing the foundation models for the image assessment task is the prospect to have a unified score, allowing for the evaluation of both technical quality and aesthetic value of an image in a single analysis. Unlike many contemporary research endeavors, our approach does not rely on predefined rules but instead uses a new data-driven method for non-parametric evaluation of image attributes. In the suggested paradigm, technical, content-based, and perceptual image qualities play key roles in shaping subjective opinions. This multi-attribute view can potentially bridge the gap between parametric and non-parametric approaches in image evaluation.

\textbf{Limitations.}
While the performance of QUASAR is promising, there are still method-related limitations that need to be considered.
First, to assess the images in terms of technical and aesthetic quality, it is necessary to build anchors for both the \textit{good} and the \textit{poor} images, which could be computationally demanding when dealing with a large image dataset.
Next, despite deteriorating the quality, the use of prompts such as “\textit{Good image.}” and “\textit{Bad image.}” is in some sense more stable because there is no way to incorrectly define these anchors. 
We stress that a commonplace issue of subjectivity in the aesthetic descriptors still applies to our study as well, which is why the values of all IAA metrics correlate with the MOS moderately.
On the other hand, even though reaching high quality, QUASAR is dependent on how we choose an image set to form the anchors. A bias in the dataset collection and/or selection may lead to poor quality of the embeddings, resulting in an inferior performance of QUASAR. A benchmarking effort and surveying of the aesthetic anchoring dataset will resolve these issues.

Improving aggregation results by carefully selecting highly informative embeddings also stands as a very promising direction for future work. Enhancing basic strategies for selecting and aggregating embeddings could involve assigning weights based on factors, such as the mean opinion score or the other characteristics during the aggregation process.


%% file: ECCV/chapters/10_conclusion.tex
\section{Conclusion}
\label{sec:conclusion}

This paper proposes QUASAR -- a data-driven framework that addresses the long-standing problem of generalizability in image quality and aesthetic assessment by offering a new robust and easily customizable, prompt-free, non-parametric metric. As such, QUASAR achieves significant improvements over the previous CLIP-based methods. 
Our work not only contributes a new and effective solution to the unified image quality and aesthetic assessments but also provides valuable insights into their inherent constituents, paving the way towards a universal technique for evaluating visual content.
 

%% file: ECCV/chapters/_supplementary.tex



\input{ECCV/chapters/12_appendix}



%% file: ECCV/chapters/12_appendix.tex

\section*{Supplementary Material}

In this Supplementary material, we show how the images could be relatively sorted according to their quality (Fig. \ref{sup:rank_iqa}) and aesthetic (Fig. \ref{sup:rank_iaa}) scores, both of which could be evaluated by the proposed unified QUASAR metric. We provide examples of good and bad quality (Figs.
\ref{sup:scores_liveitw} and 
\ref{sup:scores_spaq}) and aesthetics (Fig. \ref{sup:scores_tad}) images, with their corresponding scores, highlighting the differences in the metrics' sensitivity to various distortions and their strength (Figs. 
\ref{sup:kadid10k_distortion} and 
\ref{sup:tid2013_distortion}). 
We make special emphasis on the difference between QUASAR (ours) and CLIP-IQA (current SOTA), showing a major advantage of our image score in a variety of applications, generalizable for different sources of data. 
\input{ECCV/figs/supp_2_ranks}
\input{ECCV/figs/supp_4_ranks} 
\input{ECCV/figs/supp_1_tops}
\input{ECCV/figs/supp_3_examples}


%% file: ECCV/figs/supp_2_ranks.tex
\begin{figure*}[h]
    \centering
    \includegraphics[width=0.75\linewidth]{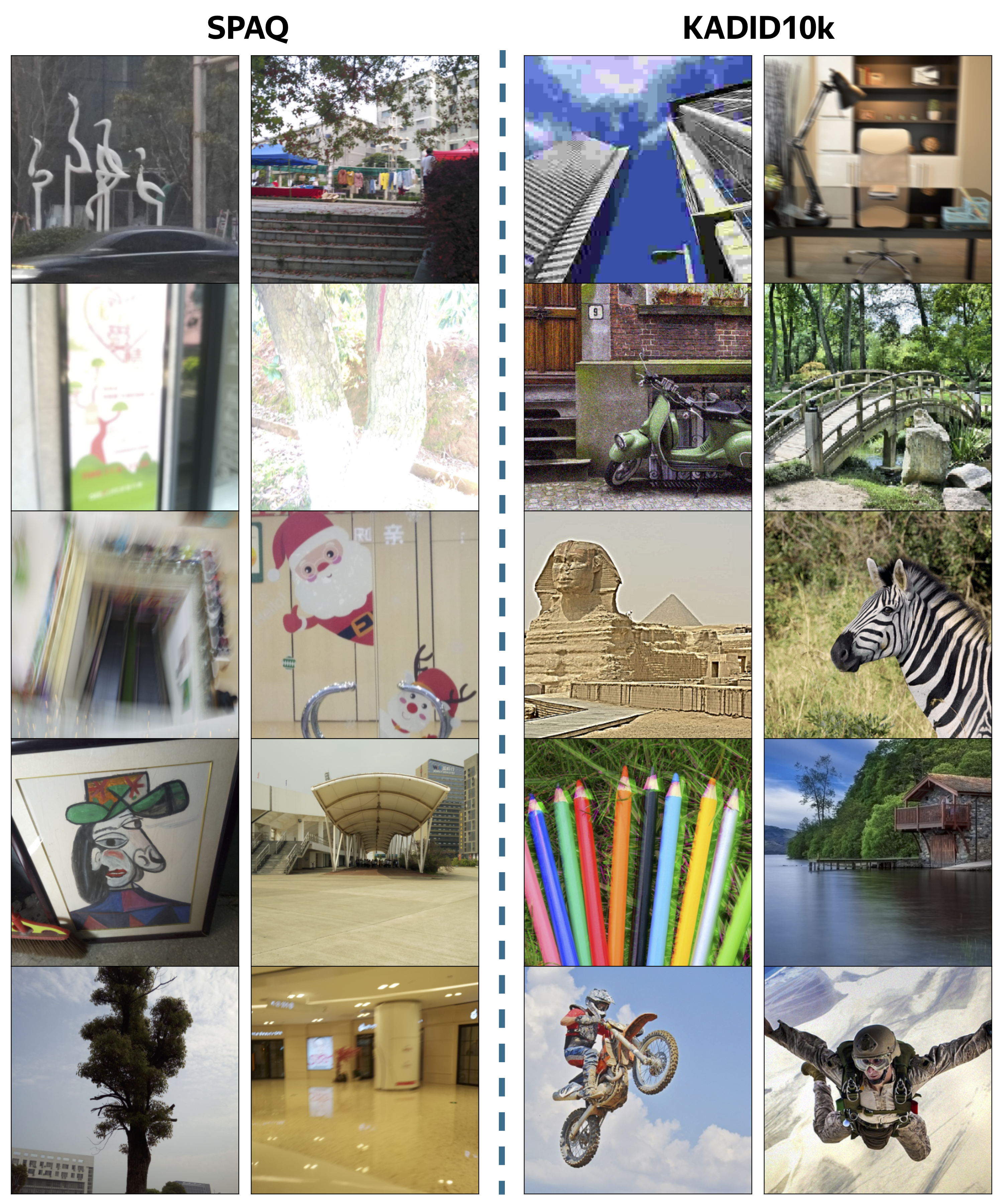}
    \caption{Example images from IQA datasets (SPAQ and KADID10k), sorted according to QUASAR (left column) and CLIP-IQA (right column) metrics. 
    The rows represent the same position in the dataset sorted according to the metrics value.
    Thus, the first row contains the worst and the last row contains the best samples from the datasets according to each metric. 
    Non-square images are center-cropped for convenience. }
    \label{sup:rank_iqa}
\end{figure*}
 

%% file: ECCV/figs/supp_4_ranks.tex
\begin{figure}[tp]
    \centering
    \includegraphics[width=\linewidth]{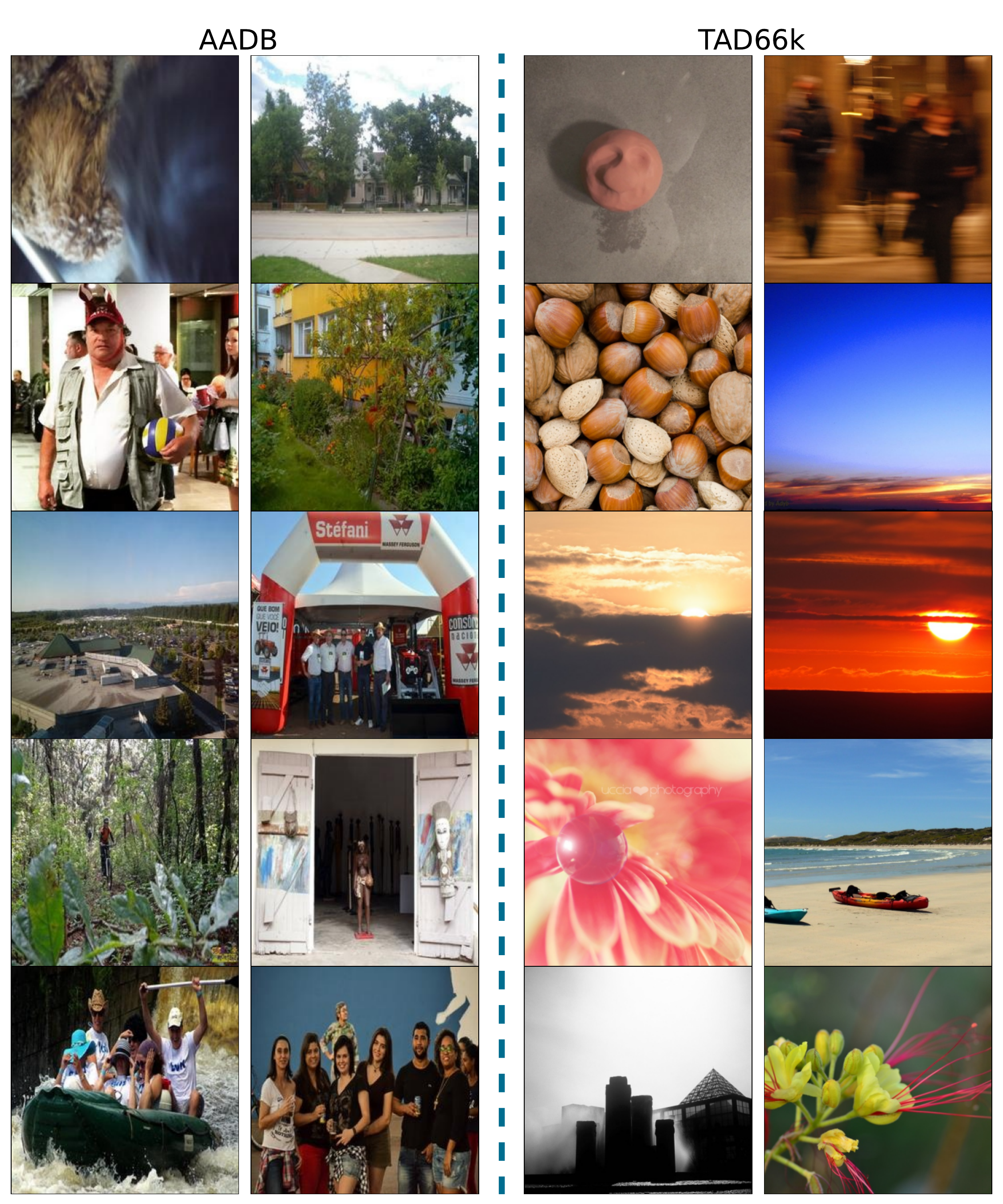}
    \caption{Examples drawn from IAA datasets (AADB and TAD66k), sorted according to QUASAR (left column) and CLIP-IQA (right column) metrics. The rows represent the same position in the dataset, sorted according to the metrics value. Thus, the first row contains the worst and the last row contains the best samples from the datasets according to each metric.}
    \label{sup:rank_iaa}
\end{figure}
 

%% file: ECCV/figs/supp_1_tops.tex
    

\begin{figure*}[tp]
    \centering
    \centering
    \begin{subfigure}{0.995\textwidth}
        \includegraphics[width=\textwidth]{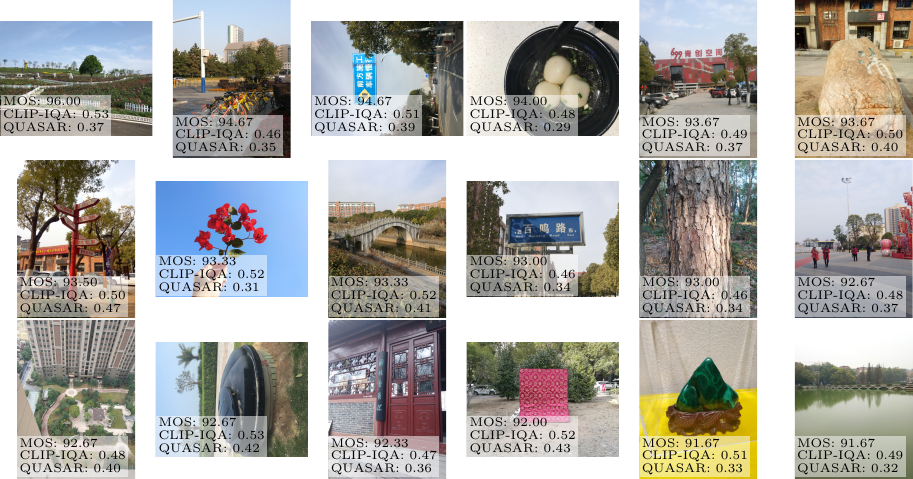}
        \caption{}
    \end{subfigure}
    \\
    \begin{subfigure}{0.995\textwidth}
        \includegraphics[width=\textwidth]{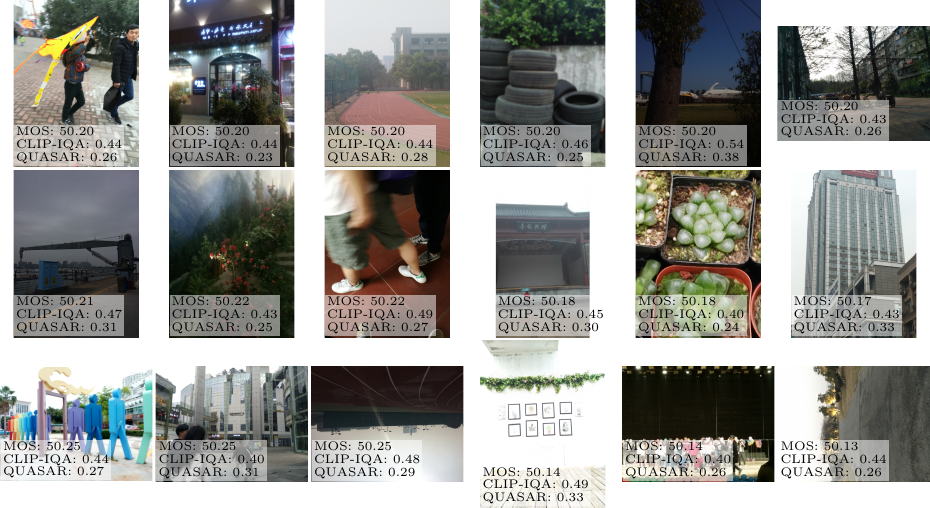}
        \caption{}
    \end{subfigure}
    \\
    \begin{subfigure}{0.995\textwidth}
        \includegraphics[width=\textwidth]{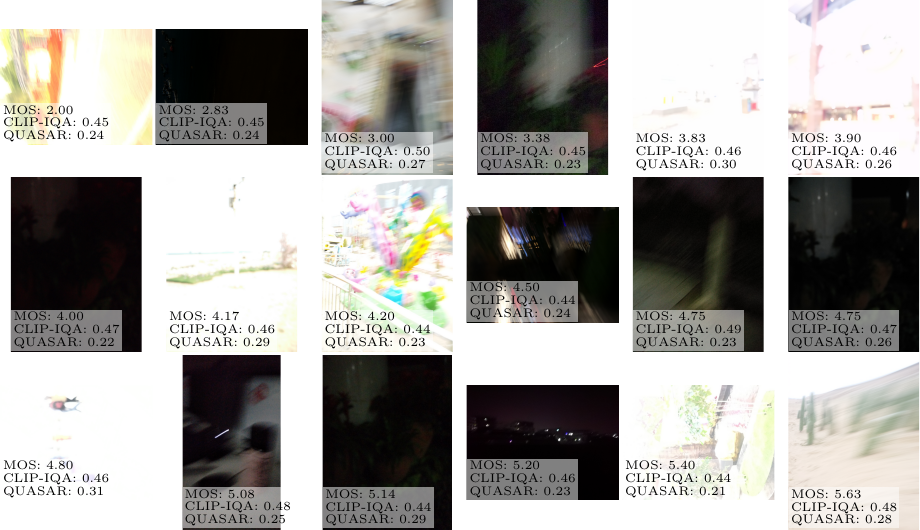}
        \caption{}
    \end{subfigure}
    
    \caption{Samples from SPAQ dataset, divided into three categories, according to MOS: (a) top, (b) median, and (c) low.}
    \label{sup:scores_spaq}
\end{figure*}

\begin{figure*}[tp]
    \centering
    \centering
    \begin{subfigure}{0.995\textwidth}
        \includegraphics[width=\textwidth]{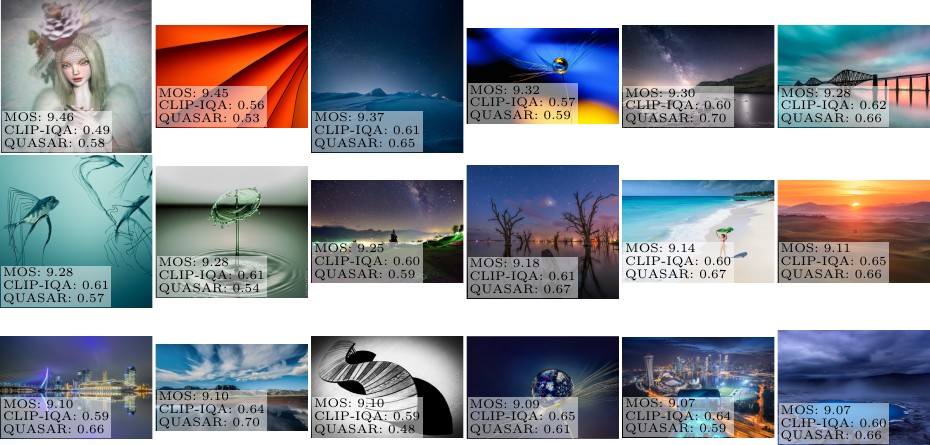}
        \caption{}
    \end{subfigure}
    \\
    \begin{subfigure}{0.995\textwidth}
        \includegraphics[width=\textwidth]{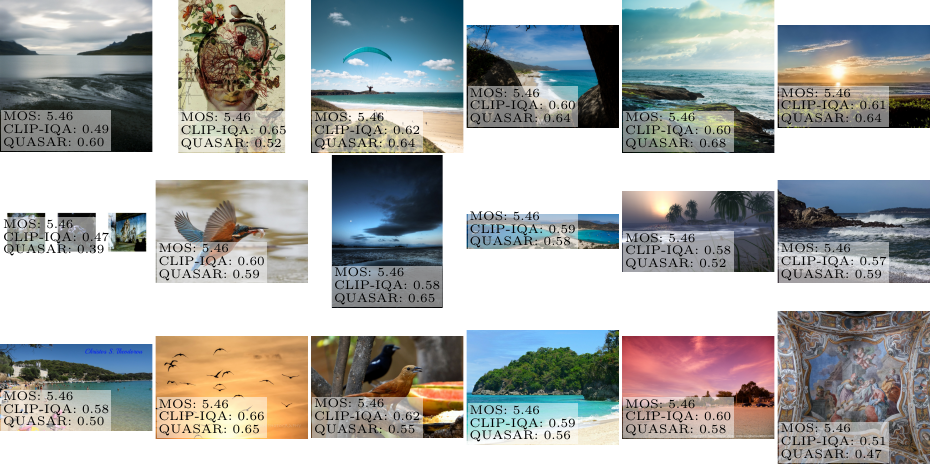}
        \caption{}
    \end{subfigure}
    \\
    \begin{subfigure}{0.995\textwidth}
        \includegraphics[width=\textwidth]{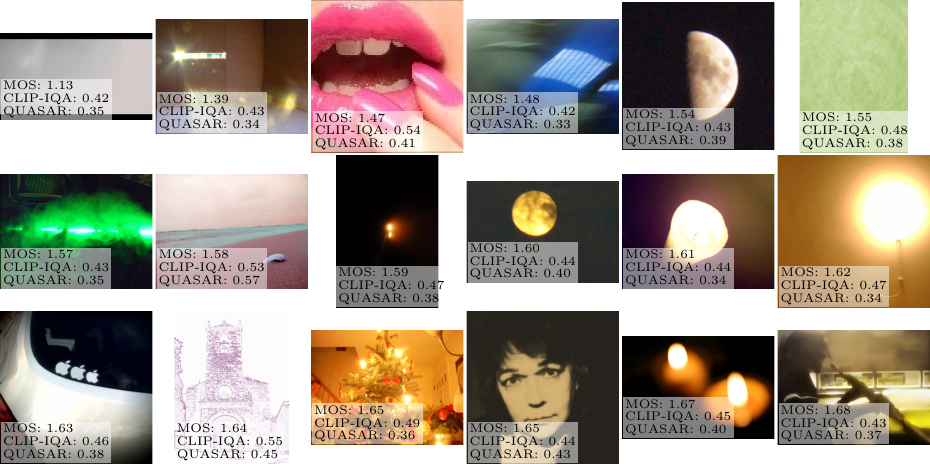}
        \caption{}
    \end{subfigure}
    
    \caption{Samples from TAD66k dataset, divided into three categories, according to MOS: (a) top, (b) median, and (c) low.}
    \label{sup:scores_tad}
\end{figure*}

%% file: ECCV/figs/supp_3_examples.tex
\begin{figure*}[tp]
    \centering
    \begin{subfigure}{0.495\textwidth}
        \includegraphics[width=\textwidth]{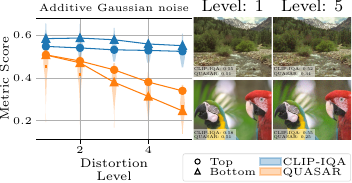}
        \caption{}
    \end{subfigure}
    \hfill
    \begin{subfigure}{0.495\textwidth}
        \includegraphics[width=\textwidth]{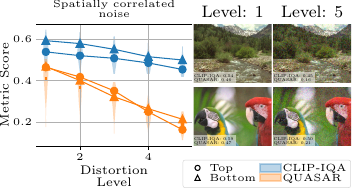}
        \caption{}
    \end{subfigure}
    \\
    \begin{subfigure}{0.495\textwidth}
        \includegraphics[width=\textwidth]{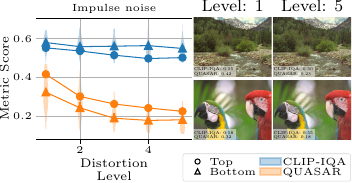}
        \caption{}

    \end{subfigure}
    \hfill
    \begin{subfigure}{0.495\textwidth}
        \includegraphics[width=\textwidth]{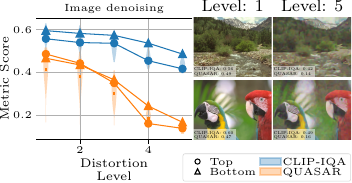}
        \caption{}

    \end{subfigure}
    \\
    \begin{subfigure}{0.495\textwidth}
        \includegraphics[width=\textwidth]{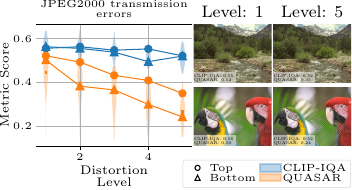}
        \caption{}

    \end{subfigure}
    \hfill
    \begin{subfigure}{0.495\textwidth}
        \includegraphics[width=\textwidth]{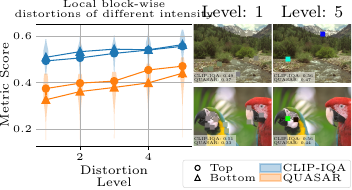}
        \caption{}

    \end{subfigure}
    \\
    \begin{subfigure}{0.495\textwidth}
        \includegraphics[width=\textwidth]{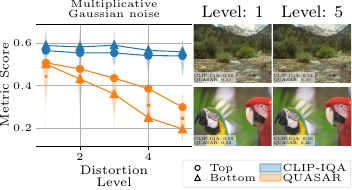}
        \caption{}

    \end{subfigure}
    \hfill
    \begin{subfigure}{0.495\textwidth}
        \includegraphics[width=\textwidth]{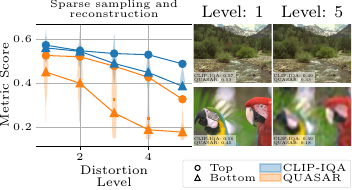}
        \caption{}

    \end{subfigure}

    \caption{Comparison of QUASAR and CLIP-IQA scores on samples from TID2013 dataset for different levels of applied distortions: (a) Additive Gaussian noise, (b) Spatially correlated noise, (c) Impulse noise, (d) Image denoising, (e) JPEG2000 transmission errors, (f) Local block-wise distortion, (g) Multiplicative Gaussian noise, and (h) Sparse sampling and reconstruction. Notice that the general trends of QUASAR and CLIP-IQA match across distortion types and levels, but QUASAR scores have a higher dynamic range. Raw images are shown, while both QUASAR and CLIP-IQA take advantage of data precprocessing, as described in the main text. The evaluated scores are normalised to $[0,1]$ range.}
    \label{sup:tid2013_distortion}
\end{figure*}